\documentclass[12pt,a4paper]{article}
\usepackage[total={7in, 10in}]{geometry}

\usepackage{graphicx}
\usepackage{helvet}
\usepackage{authblk}
\usepackage{hyperref}
\usepackage{amsmath} 
\usepackage{amssymb} 
\usepackage{orcidlink} 
\usepackage[super,comma,sort&compress]  
   {natbib}
\usepackage[right,mathlines]{lineno}
\usepackage[title]{appendix}%
\usepackage[utf8]{inputenc} 
\usepackage[T1]{fontenc}    
\usepackage{url}            
\usepackage{booktabs}       
\usepackage{amsfonts}       
\usepackage{nicefrac}       
\usepackage{adjustbox}
\usepackage{longtable}
\usepackage{microtype}      
\usepackage{xcolor}         
\usepackage{multirow}
\usepackage{threeparttable}
\usepackage{subcaption}
\usepackage{makecell}
\usepackage{placeins}
\usepackage{siunitx}     
\usepackage{caption}     
\usepackage{tabularx}
\usepackage{listings}   
\usepackage{pifont}
\usepackage{array}
\usepackage{colortbl}
\usepackage[capitalize]{cleveref}

\usepackage{algorithmic}
\usepackage{bbm}
\usepackage{textcomp}
\usepackage{algorithm}
\usepackage{xspace}
\usepackage[symbol]{footmisc}
\usepackage{scrextend}

\usepackage{rotating}

\def\BibTeX{{\rm B\kern-.05em{\sc i\kern-.025em b}\kern-.08em T\kern-.1667em\lower.7ex\hbox{E}\kern-.125emX}}
\theadset{\bfseries}

\newcommand{\rev}[1]{{\color{black}#1}}

\makeatletter
\renewcommand{\maketitle}{\bgroup\setlength{\parindent}{0pt}
\begin{flushleft}
  \textbf{\@title}
  
  \@author
\end{flushleft}\egroup}
\makeatother

\title{A \rev{Scoping} Review of Deep Learning Methods for Photoplethysmography Data}
\date{}

\author[ ]{
Guangkun Nie\textsuperscript{1,2}, 
Jiabao Zhu\textsuperscript{2,3}, 
Gongzheng Tang\textsuperscript{2,3},
Deyun Zhang\textsuperscript{4},
Shijia Geng\textsuperscript{4},
Qinghao Zhao\textsuperscript{5},
Shenda Hong\textsuperscript{2,3,*}
}

\affil[1]{School of Intelligence Science and Technology, Peking University, Beijing, China}
\affil[2]{National Institute of Health Data Science, Peking University, Beijing, China}
\affil[3]{Institute of Medical Technology, Health Science Center of Peking University, Beijing, China}
\affil[4]{HeartVoice Medical Technology, Hefei, China}
\affil[5]{Department of Cardiology, Peking University People's Hospital, Beijing, China}

\begin{document}
\linenumbers

\maketitle

\section*{ABSTRACT}

\textbf{Background:} Photoplethysmography (PPG) is a non-invasive optical sensing technique widely used to capture hemodynamic information, with broad deployment in both clinical monitoring systems and wearable devices. In recent years, the integration of deep learning has substantially advanced PPG signal analysis and expanded its applications across healthcare and non-healthcare domains.

\textbf{Methods:} We conducted a comprehensive literature search for studies applying deep learning to PPG data published between January 1, 2017 and December 31, 2025, using Google Scholar, PubMed, and Dimensions. The included studies were analyzed from three key perspectives: tasks, models, and data.

\textbf{Results:} A total of 460 papers applying deep learning techniques to PPG signal analysis were included. These studies span a wide range of application domains, from traditional physiological monitoring tasks such as cardiovascular assessment to emerging applications including sleep analysis, cross-modality signal reconstruction, and biometric identification.

\textbf{Conclusions:} Deep learning has significantly advanced PPG signal analysis by enabling more effective extraction of physiological information. Compared with traditional machine learning approaches reliant on handcrafted features, deep learning methods generally achieve improved performance and offer greater flexibility in model development. Nevertheless, several challenges remain, including limited availability of large-scale high-quality datasets, insufficient validation in real-world environments, and concerns over model interpretability, scalability, and computational efficiency. Addressing these challenges and exploring emerging research directions will be essential for further progress in deep learning-based PPG analysis.

\section*{INTRODUCTION}
In recent years, photoplethysmography (PPG) has emerged as a major sensing modality for monitoring human hemodynamics \cite{williams2023wearable, min2025wearable, petek2023consumer}. Owing to its non-invasive nature, ease of use, and suitability for continuous monitoring, PPG has been widely adopted across both clinical monitoring systems and consumer wearable devices, ranging from bedside patient monitors to smartwatches. These systems enable convenient and continuous measurement of key physiological parameters, such as heart rate (HR) \cite{pankaj2022review} and blood oxygen saturation (SpO$_2$) \cite{nitzan2014pulse}. As the deployment of PPG-enabled devices continues to expand, the volume of continuously collected signals has increased rapidly, creating a growing need for automated and reliable methods to process PPG data at scale and extract clinically and physiologically meaningful information.

Traditional machine learning approaches for automated PPG analysis typically rely on manual feature engineering. In this paradigm, hand-crafted features such as morphological, temporal, and frequency-domain characteristics are designed and extracted from PPG signals \cite{el2020review, maqsood2022survey}, and subsequently used as inputs to downstream models. Although this strategy has supported many early PPG applications, it depends heavily on expert knowledge, making feature design time-consuming and labor-intensive \cite{maqsood2022survey, pereira2020photoplethysmography}. More fundamentally, hand-crafted features are inherently limited in representational capacity, as they are derived from predefined heuristics. Consequently, they may fail to capture complex, task-relevant patterns in PPG signals, particularly those involving subtle temporal dynamics and nonlinear physiological interactions, thereby constraining the performance and scalability of traditional approaches.

Recent advances in deep learning have provided a powerful alternative to traditional approaches for PPG analysis. Unlike conventional pipelines that separate feature engineering and model training, deep learning models adopt an end-to-end framework in which feature extraction and task-specific modeling are jointly optimized \cite{lecun2015deep}. This paradigm enables models to learn informative representations directly from raw PPG signals and to capture complex physiological patterns that are difficult to characterize using hand-crafted features \cite{elul2021meeting, fujisawa2019deep}. As a result, deep learning has significantly improved the effectiveness and scalability of PPG signal analysis and has supported a wide range of applications across both healthcare and non-healthcare domains.

\begin{figure}
    \centering
    \resizebox{\linewidth}{!}{
    \includegraphics{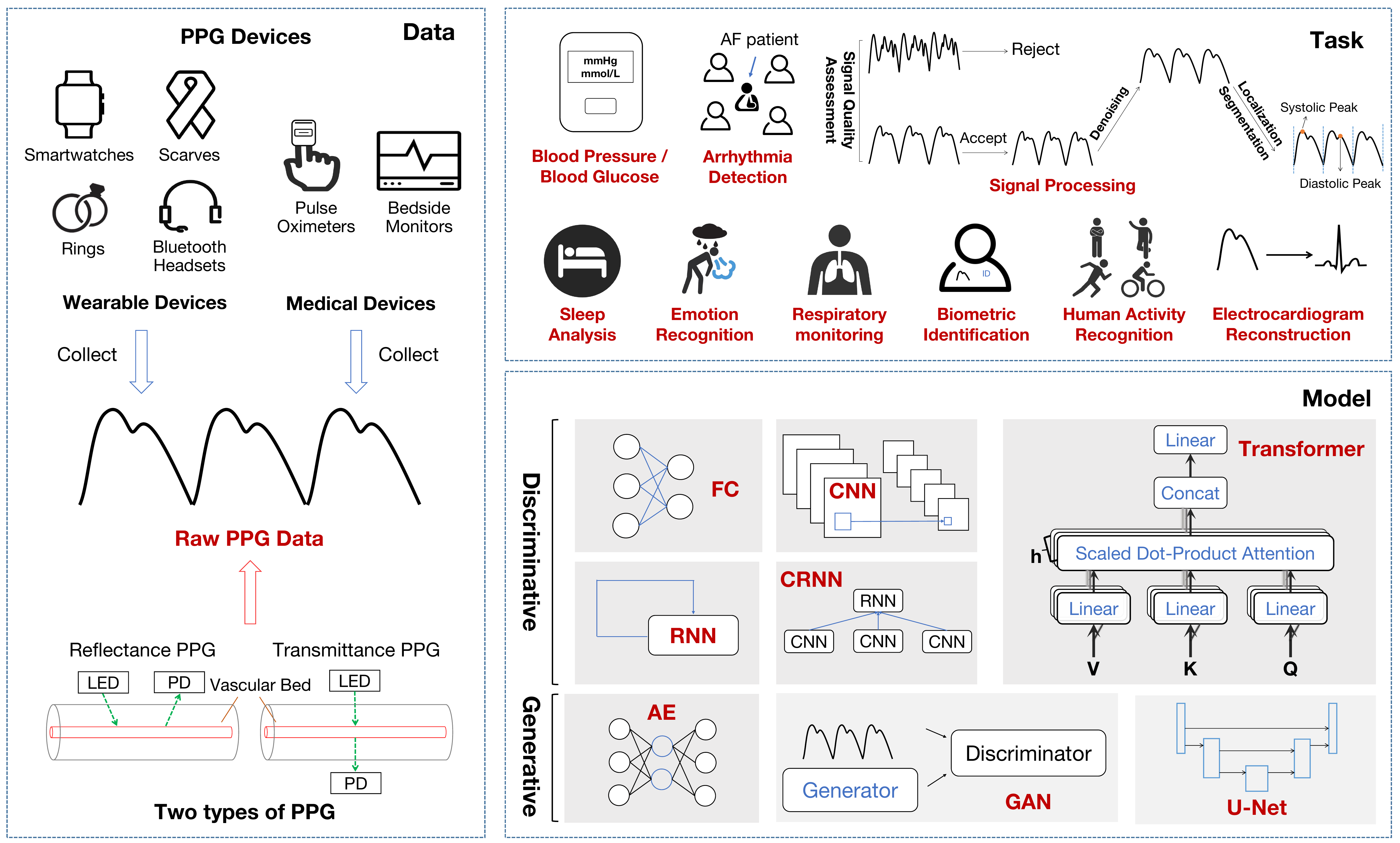}
    }
    \captionsetup{justification=raggedright,singlelinecheck=false}
    \caption{Overview of PPG signal analysis from the perspectives of tasks, models, and data. \rev{LED, light-emitting diode; PD, photodetector; FC, fully connected network; CNN, convolutional neural network; RNN, recurrent neural network; CRNN, convolutional recurrent neural network; AE, autoencoder; GAN, generative adversarial network.}}
    \label{fig:Introduction}
\end{figure}

Several previous studies have reviewed PPG-related research from different perspectives, including general healthcare applications of PPG \cite{park2022photoplethysmogram, loh2022application}, PPG-based atrial fibrillation (AF) detection \cite{pereira2020photoplethysmography}, glucose sensing using PPG \cite{jiang2025ppg}, machine learning approaches for blood pressure (BP) estimation \cite{el2020review, maqsood2022survey}, and the use of PPG in sleep monitoring \cite{ryals2023photoplethysmography}. However, a comprehensive overview focusing specifically on deep learning methods for PPG signal analysis remains limited. In this review, we provide a structured overview of deep learning approaches for PPG analysis from three perspectives: tasks, models, and data (Figure \ref{fig:Introduction}). We further highlight current challenges and emerging opportunities in this field, aiming to inform future research and practical deployment.

\section*{Methods}
\rev{This study was designed as a scoping review to provide an overview of deep learning applications for PPG data.} We conducted a literature search to identify relevant studies published between January 1, 2017 and December 31, 2025. The search was performed in the title and abstract fields across three databases: Google Scholar\footnote{https://scholar.google.com/}, PubMed\footnote{https://pubmed.ncbi.nlm.nih.gov/}, and Dimensions\footnote{https://www.dimensions.ai/}. The search query used in all databases was ("deep learning" OR "DL") AND ("photoplethysmography" OR "PPG"), with all keywords treated as case-insensitive.

\begin{figure}
    \centering
    \includegraphics[scale=0.5]{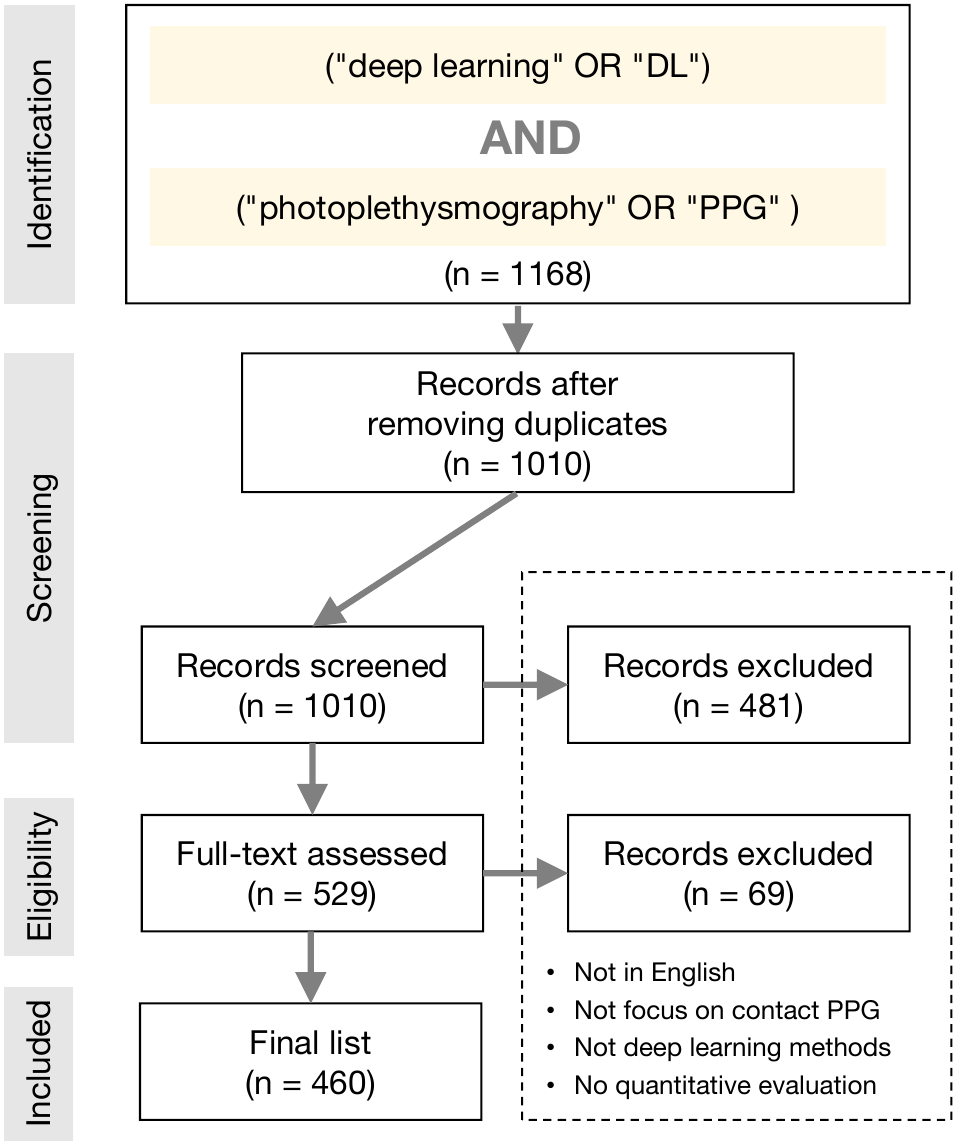}
    \caption{Workflow of the literature search and selection process.}
    \label{fig:workflow}
\end{figure}

The literature search and selection workflow is illustrated in Figure \ref{fig:workflow}. The process followed four stages: identification, screening, eligibility, and inclusion. Initially, 1,168 papers were retrieved, which were reduced to 1,010 after duplicate removal. The titles and abstracts of these papers were then screened according to the following exclusion criteria: 1) not written in English; 2) not focusing on contact PPG data; 3) not focusing on deep learning methods; and 4) lacking quantitative evaluation. Based on these criteria, 481 papers were excluded during the screening stage. The remaining papers were subsequently assessed through full-text review, resulting in the exclusion of an additional 69 papers. Ultimately, 460 papers were included in the final analysis.

\section*{Results}
A total of 460 papers were included in this review and analyzed from three key perspectives: tasks, models, and data. Based on their research objectives, the reviewed studies were categorized into eleven task groups, including BP analysis (N=154), cardiovascular monitoring and diagnosis (N=106), sleep health (N=27), mental health (N=31), respiratory monitoring and analysis (N=26), blood glucose analysis (N=31), signal processing (N=41), biometric identification (N=27), electrocardiogram (ECG) reconstruction (N=16), human activity recognition (N=7), and others (N=23). Some papers addressed multiple tasks and were therefore counted in more than one category. In addition, studies related to BP and blood glucose analysis were reported separately from cardiovascular monitoring and diagnosis due to their relatively large number. Table \ref{data_models} summarizes the tasks, datasets, and models used in a representative subset of the reviewed studies, while a detailed summary of included papers is available in our GitHub repository at \url{https://github.com/Ngk03/DL_PPG_Review}.

\begin{scriptsize}
\captionsetup{justification=raggedright,singlelinecheck=false}
\begin{longtable}{>{\raggedright\arraybackslash}m{3.5cm}
                  >{\raggedright\arraybackslash}m{4cm}
                  >{\raggedright\arraybackslash}m{4.5cm}
                  >{\raggedright\arraybackslash}m{4.2cm}}

\caption{Summary of representative studies on deep learning-based PPG analysis. VPG, velocity plethysmogram; APG, acceleration plethysmogram; ECG, electrocardiogram; BP, blood pressure; ABP, arterial blood pressure; CWT, continuous wavelet transform; STFT, short-time Fourier transform; CRNN, convolutional recurrent neural network; GRU, gated recurrent unit; LSTM, long short-term memory; CNN, convolutional neural network; RNN, recurrent neural network; FC, fully connected network; AE, autoencoder; GAN, generative adversarial network.}
\label{data_models} \\

\hline
\textbf{Task} & \textbf{Citation} & \textbf{Input Data} & \textbf{Model} \\
\hline
\endfirsthead

\multicolumn{4}{c}{\textit{Table \thetable\ continued from previous page}}\\
\hline
\textbf{Task} & \textbf{Citation} & \textbf{Input Data} & \textbf{Model} \\
\hline
\endhead

\hline
\multicolumn{4}{r}{\textit{Continued on next page}}\\
\hline
\endfoot

\hline
\endlastfoot
Human activity recognition &
  Boukhechba et al. (2019) \cite{boukhechba2019actippg_27} & PPG-derived frequency components (cardiac, respiratory, and motion-related) & CRNN (CNN-LSTM) \\
Human activity recognition &
  Mekruksavanich et al. (2022) \cite{mekruksavanich2022cnn_59} &
  PPG, accelerometer &
  CNN \\
Biometric identification &
  Biswas et al. (2019) \cite{biswas2019cornet_63} &
  PPG &
  CRNN (CNN-LSTM) \\
Blood glucose analysis &
  Chen et al. (2024) \cite{chen2024multi} &
  PPG, VPG, APG, kinetic features &
  CNN, Transformer \\
Blood glucose analysis &
  Chowdhury et al. (2024) \cite{chowdhury2024mmg} &
  PPG, electrodermal activity, skin temperature, food logs &
  CNN \\
BP analysis &
  Li et al. (2020) \cite{li2020real_168} &
  PPG and ECG features &
  LSTM \\
BP analysis &
  Cheng et al. (2021) \cite{cheng2021prediction_162} &
  PPG, VPG, APG &
  CNN, U-Net \\
BP analysis &
  Panwar et al. (2020) \cite{panwar2020pp_160} &
  PPG &
  CRNN (CNN-LSTM) \\
BP analysis &
  Leitner et al. (2021) \cite{leitner2021personalized_148} &
  PPG &
  CRNN (CNN-GRU) \\
BP analysis &
  Radha et al. (2019) \cite{radha2019estimating_105} &
  PPG features &
  FC, LSTM \\
BP analysis &
  Lee et al. (2021) \cite{lee2021deep_90} &
  PPG, ECG, ABP, capnography &
  CNN \\
BP analysis &
  El-Hajj et al. (2021) \cite{el2021deep_88} &
  PPG features &
  LSTM, GRU \\
BP analysis &
  Wang et al. (2021) \cite{wang2021cuff_66} &
  PPG (visibility graph) &
  CNN \\
BP analysis &
  Aguirre et al. (2021) \cite{aguirre2021blood_47} &
  PPG &
  GRU \\
BP analysis & Slapničar et al. (2019) \cite{slapnivcar2019blood_45} & PPG, VPG, APG, and their corresponding spectrograms (STFT) & CNN, GRU \\
BP analysis &
  Athaya et al. (2021) \cite{athaya2021estimation_30} &
  PPG &
  CNN, U-Net \\
BP analysis &
  Cheng et al. (2022) \cite{chen2022new_17} &
  PPG &
  CNN \\
BP analysis &
  Esmaelpoor et al. (2020) \cite{esmaelpoor2020multistage_15} &
  PPG &
  CRNN (CNN-LSTM) \\
BP analysis &
  Liang et al. (2018) \cite{liang2018photoplethysmography_147} &
  PPG (CWT) &
  CNN \\
BP analysis &
  Wang et al. (2024) \cite{DBLP:journals/titb/WangMKN24} &
  PPG, VPG, APG &
  CNN, CRNN (CNN-LSTM) \\
BP analysis &
  Liu et al. (2024) \cite{DBLP:journals/cbm/LiuZZ24} &
  PPG &
  GRU \\
BP analysis &
  Ma et al. (2024) \cite{ma2024upr} &
  PPG &
  Transformer \\
BP analysis &
  Ma et al. (2024) \cite{ma2024diffcnbp} &
  PPG, demographic information &
  CNN, Transformer, diffusion model \\
BP analysis &
  Liu et al. (2024) \cite{liu2024large} &
  PPG, VPG, APG, and ECG features &
  LLM \\
BP analysis &
  Chen et al. (2025) \cite{chen2025versatile} &
  PPG, ECG &
  Transformer, diffusion model \\
Mental health &
  Lee et al. (2019) \cite{lee2019fast_111} &
  PPG &
  CNN \\
Mental health &
  Choi et al. (2023) \cite{choi2023weighted_195} &
  PPG (STFT) &
  CNN, CRNN (CNN-LSTM) \\
Mental health &
  Li et al. (2025) \cite{li2025stress} &
  PPG-derived signals (pulse rate variability, beat-level PPG) &
  CRNN (CNN-LSTM) \\
Sleep health &
  Radha et al. (2021) \cite{radha2021deep_11} &
  PPG features &
  LSTM \\
Sleep health &
  Huttunen et al. (2021) \cite{huttunen2021assessment_32} &
  PPG &
  CRNN (CNN-LSTM) \\
Sleep health &
  Korkalainen et al. (2020) \cite{korkalainen2020deep_81} &
  PPG &
  CRNN (CNN-GRU) \\
Sleep health &
  Kotzen et al. (2022) \cite{kotzen2022sleepppg_185} &
  PPG &
  CNN \\
Sleep health &
  Papini et al. (2020) \cite{papini2020wearable_194} &
  PPG features &
  CNN \\
Sleep health &
  Wang et al. (2025) \cite{wang2025dual} &
  PPG, pulse pressure wave, SpO$_2$ &
  Transformer \\
ECG reconstruction &
  Sarkar et al. (2021) \cite{sarkar2021cardiogan_51} &
  PPG &
  CNN, U-Net, GAN \\
ECG reconstruction &
  Chiu et al. (2020) \cite{chiu2020reconstructing_171} &
  PPG &
  Transformer, CNN \\
ECG reconstruction &
  Vo et al. (2021) \cite{vo2021p2e_143} &
  PPG &
  CNN, U-Net, GAN \\
ECG reconstruction &
  Ding et al. (2025) \cite{ding2025ai} &
  PPG &
  Transformer \\
ECG reconstruction &
  Chen et al. (2025) \cite{chen2025versatile} &
  PPG, ABP &
  Transformer, diffusion model \\
Respiratory monitoring &
  Ravichandran et al. (2019) \cite{ravichandran2019respnet_175} &
  PPG &
  CNN, U-Net \\
Respiratory monitoring &
  Bian et al. (2020) \cite{bian2020respiratory_174} &
  PPG &
  CNN \\
Respiratory monitoring &
  Kumar et al. (2022) \cite{kumar2022deep_83} &
  PPG, ECG, and electromyography &
  CNN, LSTM, CRNN (CNN-LSTM) \\
Cardiovascular monitoring &
  Shashikumar et al. (2017) \cite{shashikumar2017deep_7} &
  PPG (CWT), PPG features &
  CNN \\
Cardiovascular monitoring &
  Shen et al. (2019) \cite{shen2019ambulatory_28} &
  PPG &
  CNN \\
Cardiovascular monitoring &
  Kwon et al. (2019) \cite{kwon2019deep_77} &
  PPG &
  CNN, RNN \\
Cardiovascular monitoring &
  Allen et al. (2021) \cite{allen2021deep_92} &
  PPG (CWT) &
  CNN \\
Cardiovascular monitoring &
  Reiss et al. (2019) \cite{reiss2019deep_93} &
  PPG (CWT) &
  CNN \\
Cardiovascular monitoring &
  Torres-Soto et al. (2020) \cite{torres2020multi_136} &
  PPG &
  CNN \\
Cardiovascular monitoring &
  Weng et al. (2025) \cite{weng2024predicting} &
  PPG, age, sex, and smoking status &
  CNN \\
Cardiovascular monitoring &
  Ding et al. (2025) \cite{ding2025ai} &
  PPG &
  Transformer \\
Cardiovascular monitoring &
  Wang et al. (2025) \cite{wang2025thinking} &
  PPG &
  CNN \\
Cardiovascular monitoring &
  Nie et al. (2025) \cite{nie2025artificial} &
  PPG &
  CNN \\
Cardiovascular monitoring &
  Miller et al. (2025) \cite{miller2025wearable} &
  PPG &
  CNN \\
Signal processing &
  Xu et al. (2020) \cite{xu2020stochastic_186} &
  PPG &
  LSTM \\
Signal processing &
  Goh et al. (2020) \cite{goh2020robust_180} &
  PPG &
  CNN \\
Signal processing &
  Pereira et al. (2019) \cite{pereira2019deep_76} &
  PPG &
  CNN \\
Signal processing &
  Shin et al. (2022) \cite{shin2022deep_74} &
  PPG &
  CNN \\
Signal processing &
  Ding et al. (2023) \cite{ding2023log_128} &
  PPG &
  CNN, GAN \\
Signal processing &
  Zheng et al. (2024) \cite{zheng2024tiny} &
  PPG &
  CNN \\
Signal processing &
  Mohagheghian et al. (2024) \cite{mohagheghian2024atrial} &
  PPG &
  CNN, AE \\
Others &
  Tadesse et al. (2020) \cite{tadesse2020multi_135} &
  PPG (STFT), ECG spectrograms (STFT) &
  LSTM, CRNN (CNN-LSTM) \\
Others &
  Chowdhury et al. (2021) \cite{chowdhury2021deep_91} &
  PPG, ECG &
  CNN \\
Others &
  Lombardi et al. (2022) \cite{lombardi2022classifying_57} &
  PPG &
  CNN \\
Others &
  Lyle et al. (2025) \cite{lyle2025artificial} &
  PPG &
  Transformer \\
Others &
  Karolcik et al. (2024) \cite{karolcik2024towards} &
  PPG &
  CNN \\
Others &
  Yu et al. (2023) \cite{yu2023artificial} &
  PPG &
  CNN, CRNN (CNN-LSTM), AE \\
  \hline
\end{longtable}
\end{scriptsize}

\subsection*{Tasks}
\subsubsection*{Blood pressure analysis}
BP is a critical cardiovascular indicator, and sustained abnormalities substantially increase the risk of adverse outcomes \cite{verdecchia20202020}. As a non-invasive and wearable-compatible modality, PPG has been extensively investigated for cuffless BP assessment, since clinically established invasive arterial lines and intermittent cuff-based measurements are impractical for unobtrusive and continuous monitoring in daily settings.

Existing studies on BP analysis mainly address two objectives: BP classification (e.g., hypertension detection) and BP value estimation (e.g., systolic blood pressure and diastolic blood pressure prediction). In BP classification, a given PPG signal is mapped to predefined BP categories. Depending on the diagnostic granularity, this task has been formulated as binary classification (e.g., normotension vs. hypertension) \cite{yan2023photoplethysmography_152, wu2021improving_123, liang2018photoplethysmography_147}, three-class classification (normotension, prehypertension, and hypertension) \cite{kumar2023optimized_142, cano2021hypertension_122}, or finer-grained multi-class settings that further distinguish stage 1 and stage 2 hypertension, and occasionally hypotension \cite{yen2021deep_75}. Beyond categorical prediction, most studies focus on BP estimation, which can be broadly divided into two directions. The first predicts discrete BP metrics from PPG signals, including systolic blood pressure, diastolic blood pressure, and mean arterial BP \cite{slapnivcar2019blood_45, panwar2020pp_160, leitner2021personalized_148, li2020real_168}. The second reconstructs continuous arterial blood pressure (ABP) waveforms \cite{cheng2021prediction_162, kim2022deepcnap_95}, enabling more comprehensive hemodynamic characterization. These approaches are typically formulated as sequence-to-sequence learning problems and implemented using architectures such as U-Net variants \cite{cheng2021prediction_162}, generative adversarial network (GAN) variants \cite{mehrabadi2022novel_141}, and more recently diffusion models \cite{DBLP:journals/iotj/MaGZLZ25}. 

More recent studies further shift attention from absolute BP values within short time windows to relative BP variations and long-term trends. For example, $\Delta$BP-Net employs a self-contrastive masking strategy to model temporal BP changes \cite{DBLP:journals/titb/WangMKN24}. Other approaches combine convolutional neural network (CNN) with Transformer \cite{huang2025robust} or gated recurrent unit (GRU) with attention \cite{DBLP:journals/cbm/LiuZZ24} to capture long-term BP dynamics. These studies highlight an emerging trend toward modeling the temporal evolution of BP rather than relying solely on instantaneous estimates.

\subsubsection*{Cardiovascular monitoring and diagnosis}
PPG signals encode rich hemodynamic information and are therefore widely used in cardiovascular monitoring and analysis. From a functional perspective, existing applications can be broadly grouped into three categories: physiological parameter estimation, clinical condition identification, and incident disease risk modeling.

At the most fundamental level, PPG is used to estimate cardiovascular parameters that reflect an individual's physiological state. Representative tasks include HR estimation and heart rate variability analysis \cite{biswas2019cornet_63, reiss2019deep_93}. Beyond these commonly studied indicators, additional hemodynamic indices have also been explored, such as pulse wave velocity \cite{abrisham2024advancing}, arterial stiffness \cite{abrisham2024deep}, systemic vascular resistance \cite{ipar2025parallel}, and cardiac output \cite{xu2023improved, zhao2025cardiac}. These tasks aim to derive quantitative physiological variables from PPG signals, providing a foundation for higher-level clinical interpretation.

Beyond physiological parameter estimation, PPG has been extensively employed for cardiovascular disease detection and diagnosis. Representative applications include arrhythmia detection and classification, such as premature ventricular contraction, premature atrial contraction, ventricular tachycardia, supraventricular tachycardia, and atrial fibrillation \cite{liu2022multiclass_132, torres2020multi_136, antiperovitch2024continuous}. Other studies have investigated myocardial infarction detection \cite{gomez20251d}, heart failure identification \cite{tiosavljevic2025heart}, coronary artery disease detection \cite{sivasubramaniam2025early, minhas2025machine}, and localization of stenotic coronary arteries \cite{md2024deep}. In addition, recent work has demonstrated the feasibility of using consumer-grade smartwatch PPG recordings to detect out-of-hospital cardiac arrest, highlighting the potential of wearable PPG for real-world continuous monitoring of critical cardiovascular events \cite{shah2025automated}.

Beyond disease recognition, research has further expanded toward risk stratification and long-term outcome modeling. For example, Weng et al. \cite{weng2024predicting} developed deep learning-derived PPG representations to predict the 10-year risk of major adverse cardiovascular events. In parallel, studies on biological aging have introduced PPG age estimation frameworks \cite{nie2025artificial, miller2025wearable}, in which complex PPG signals are summarized into scalar indices reflecting cardiovascular aging status and demonstrating associations with future adverse cardiovascular events.

\subsubsection*{Sleep health}
Sleep health is closely linked to overall physiological and mental well-being. Polysomnography (PSG), the clinical gold standard for sleep assessment, is costly, labor-intensive, and may interfere with natural sleep due to its intrusive setup \cite{bruyneel2011sleep}. In contrast, PPG provides a scalable and wearable-friendly alternative for large-scale and longitudinal sleep monitoring.

Sleep staging represents one of the most extensively studied tasks. Korkalainen et al. \cite{korkalainen2020deep_81} proposed a convolutional recurrent neural network (CRNN)-based architecture for sleep stage classification in patients with suspected sleep apnea using PPG-derived signals. To further improve robustness and generalization, several studies adopted pretraining and transfer learning strategies. For example, models pretrained on ECG signals have been successfully transferred to PPG-based sleep staging tasks, achieving improved accuracy and cross-dataset generalization compared with single-stage end-to-end training \cite{radha2021deep_11, li2021transfer_192, kotzen2022sleepppg}. Extending this idea, Haimov et al. \cite{haimov2024deep} applied transfer learning to adapt adult-trained sleep staging models to pediatric populations, highlighting the flexibility of representation transfer in PPG-based sleep analysis.

Beyond sleep staging, PPG has also been widely explored for detecting sleep-disordered breathing, particularly sleep apnea. Most approaches formulate apnea detection as a time-window-level classification problem. Papini et al. \cite{papini2020wearable_194} employed a CNN to detect apnea events within 30-second windows using 212 feature sequences extracted from wrist-worn reflective PPG. Wei et al. \cite{wei2023ms_131} leveraged PPG-derived peak-to-peak interval and amplitude decrease sequences to identify apnea events at a one-minute resolution. Moving toward end-to-end modeling, Wang et al. \cite{wang2025dual} developed a dual-modal wearable pulse detection system that applies Transformer-based modeling to raw PPG, pulse wave pressure, and SpO$_2$ signals for apnea event detection. In addition to apnea detection, PPG-based models have also been explored for other sleep-related tasks, including insomnia identification \cite{telangore2025ppg}, nocturnal epileptic seizure detection \cite{dong2025detection}, and estimation of sleep metrics such as arousal index and total sleep time \cite{li2024cross}.

\subsubsection*{Mental health}
Emotional and stress-related states are closely associated with autonomic nervous system activity and cardiovascular regulation. As PPG reflects peripheral hemodynamic changes and heart rate variability dynamics, it provides a non-invasive proxy for capturing emotion-induced physiological responses, making it a promising modality for affective state recognition.

Most existing studies focus on stress detection, which is typically formulated as a binary classification problem (stress vs. non-stress). Deep learning frameworks combining CRNN architectures with temporal attention and self-distillation strategies have been shown to improve robustness and classification performance in this setting \cite{choi2023weighted_195}. However, binary discrimination provides only coarse-grained information. To enable more fine-grained assessment, subsequent work has explored direct regression of stress scores based on standardized psychological scales, leveraging PPG-derived pulse rate variability features and beat-level segmented PPG sequences \cite{li2025stress}. Beyond stress-focused modeling, several studies further extend the task to multi-class emotion recognition, differentiating affective states such as baseline, stress, amusement, and meditation \cite{choi2023weighted_195, choi2022attention_35}, thereby providing a more detailed characterization of emotional states.

\subsubsection*{Respiratory monitoring and analysis}
Respiratory disorders are highly prevalent and often accompanied by alterations in breathing patterns. Conventional respiratory assessment methods, such as capnography systems and nasal/oral pressure transducers, require specialized instrumentation and are inconvenient for long-term or out-of-clinic use \cite{charlton2017breathing}. In this context, PPG-based techniques have emerged as a practical alternative for unobtrusive respiratory monitoring.

Among these applications, respiratory rate (RR) estimation is the most extensively studied task. End-to-end deep learning models have been developed to directly estimate RR from raw PPG signals. For example, a ResNet-based framework demonstrated accurate RR prediction, with further improvements achieved by incorporating synthetic PPG data to enhance robustness \cite{bian2020respiratory_174}. Sequence modeling approaches have also been explored. A bidirectional long short-term memory (LSTM) model combined with Bahdanau attention showed reliable performance in capturing respiratory dynamics \cite{kumar2022deep_83}. Beyond estimating RR as a single scalar parameter, several studies have focused on reconstructing continuous respiratory waveforms from PPG signals. Encoder-decoder architectures such as RespNet \cite{ravichandran2019respnet_175}, which employs dilated residual inception blocks, and waveform-oriented models such as RRWaveNet \cite{osathitporn2023rrwavenet_183} enable extraction of full respiratory signals. Similarly, U-Net-based mappings from PPG to capnography references have demonstrated that reconstructed waveforms can closely approximate gold-standard respiratory measurements \cite{davies2023rapid_166}.

In addition to breathing pattern analysis, PPG has also been explored for oxygenation-related tasks. Deep learning approaches have been proposed for direct SpO$_2$ estimation \cite{chu2023non_139}, as well as for hypoxemia detection and severity prediction \cite{sundrani2023predicting_161, mahmud2022res_177}.

\subsubsection*{Blood glucose analysis}
Diabetes is a prevalent metabolic disorder characterized by chronic hyperglycemia and associated systemic complications \cite{alemayehu2020knowledge, gao2025prevalence}. Conventional blood glucose monitoring relies on invasive finger-prick measurements, which are inconvenient for frequent or long-term use \cite{sarkar2018design, mekonnen2020accurate}. Because glucose concentration affects optical absorption and scattering properties at specific wavelengths \cite{paul2012design}, PPG signals have been explored as a potential non-invasive surrogate for blood glucose assessment.

Most existing studies focus on direct glucose estimation from PPG signals. For example, Chen et al. \cite{chen2024multi} proposed a Transformer-based architecture that integrates raw PPG waveforms with kinetic features through multi-view feature fusion for instantaneous glucose prediction. Other studies incorporate historical or pretest information to improve prediction stability. In Ref. \cite{chu2025improving}, a pretest PPG recording and its corresponding blood glucose value are combined with the current PPG measurement to first estimate the subject's current HbA1c level. The estimated HbA1c is then integrated with these inputs in a subsequent stage to infer the current blood glucose value. Beyond direct glucose estimation, PPG has also been used to infer related hematological indicators such as hemoglobin concentration \cite{smarandache2025noninvasive}, and to support disease-level screening tasks including diabetes detection \cite{zeynali2025non} and hypoglycemia identification in premature infants \cite{shafiq2025dual}.

\subsubsection*{Signal processing}
In addition to downstream health-related applications, deep learning has also been widely applied to PPG signal processing tasks. Since PPG signals acquired in real-world settings are often affected by motion artifacts, sensor noise, and other sources of interference, signal quality assessment has become one of the most extensively studied directions. Most approaches formulate signal quality assessment as a classification task that distinguishes clean signals from artifact-contaminated ones. For example, CNN-based models have been used to classify short PPG segments (e.g., five-second windows) as either high-quality or corrupted by motion artifacts \cite{goh2020robust_180}. To achieve finer temporal resolution, Zheng et al. \cite{zheng2024tiny} proposed Tiny-PPG, a fully convolutional architecture for point-wise artifact detection that enables real-time signal quality monitoring.

Beyond quality assessment, deep learning has also been explored for PPG denoising, artifact correction, and waveform structure analysis. Hybrid approaches combining neural networks with classical signal processing techniques have been proposed. For instance, a fully connected network (FC) integrated with wavelet-based multi-resolution analysis has been used to suppress noise in PPG recordings \cite{ahmed2023deep_4}. Generative models such as GANs and autoencoders have also been investigated to reconstruct corrupted PPG signals, thereby mitigating the impact of artifacts on downstream analysis \cite{wang2022ppg_155, mohagheghian2024atrial}. Another related direction focuses on waveform structure analysis, including fiducial point detection and signal segmentation. In Ref. \cite{marzorati2022hybrid_121}, a CNN-based model has been employed to localize key fiducial points in the PPG waveform, such as the foot, maximum slope, and systolic peak, facilitating the extraction of morphological features. In addition, recurrent architectures such as bidirectional LSTM have been used to segment PPG signals into systolic and diastolic phases, providing structured representations of the pulse waveform \cite{xu2019deep_94}.

\subsubsection*{Biometric identification}
Biometric identification based on physiological signals has attracted increasing attention as a secure alternative or complement to conventional approaches such as fingerprint and facial recognition \cite{wang2020differences}. Due to the person-specific morphological and temporal characteristics embedded in pulse waveforms, PPG signals provide a promising modality for biometric authentication.

Deep learning models have been widely explored to learn discriminative representations from PPG signals. For example, CorNET \cite{biswas2019cornet_63} employs a hybrid architecture consisting of convolutional layers for feature extraction and LSTM layers for temporal modeling, followed by a dense classifier, achieving an average identification accuracy of 96\% across 20 subjects using single-channel wrist-worn PPG signals. Similar hybrid architectures combining CNN and recurrent layers have also been investigated with different configurations to improve identification robustness \cite{hwang2020evaluation_109, liu2021deep_84, xiong2021stable_23}. Beyond modeling raw time-series signals, several studies have explored alternative signal representations to enhance identity discrimination. CNN-based frameworks have been applied not only to raw PPG sequences \cite{wang2022biometric_39}, but also to image-like representations derived from PPG signals, such as Gram matrix features and spectrogram transformations \cite{wu2022gram_118, siam2021biosignal_41}.

\subsubsection*{Electrocardiogram reconstruction}
Reconstructing ECG signals from PPG recordings has attracted growing interest in recent years. This direction is motivated by the intrinsic physiological coupling between cardiac electrical activity and the peripheral hemodynamic responses reflected in PPG signals \cite{chiu2020reconstructing_171}. While PPG sensors are widely integrated into wearable devices due to their simplicity and comfort, ECG remains the clinical gold standard for diagnosing many cardiac conditions. Therefore, reconstructing ECG waveforms from PPG offers a promising approach to bridge wearable sensing and clinically informative cardiac monitoring \cite{vo2021p2e_143}.

Various deep learning frameworks have been proposed to learn the mapping from PPG to ECG signals. For example, CardioGAN \cite{sarkar2021cardiogan_51} introduced a CycleGAN-based architecture with dual discriminators operating in both temporal and spectrogram domains to improve waveform reconstruction fidelity. To better preserve clinically important ECG components, Chiu et al. \cite{chiu2020reconstructing_171} proposed a QRS complex-enhanced loss that places greater emphasis on reconstructing the QRS region. More recently, generative frameworks based on diffusion Transformer architectures have also been explored to achieve more accurate ECG waveform reconstruction \cite{chen2025versatile}. In addition to advances in generative modeling, several studies have investigated alignment strategies between PPG and ECG signals to further improve reconstruction fidelity. For instance, Tang et al. \cite{tang2022robust_182} aligned the systolic peak of the PPG waveform with the corresponding ECG R-peak prior to reconstruction to reduce temporal mismatch. Other works explored feature-level alignment during representation learning to enhance the physiological consistency between reconstructed ECG signals and their PPG counterparts \cite{vo2021p2e_143, tang2023ppg2ecgps_157, ding2025ai}.

\subsubsection*{Human activity recognition}
Human activity recognition plays an important role in applications such as healthcare monitoring and rehabilitation surveillance \cite{boukhechba2019actippg_27, mekruksavanich2022classification_55}. With the widespread integration of PPG sensors in wearable devices such as smartwatches and wristbands, increasing attention has been directed toward exploring their potential for activity recognition.

One approach leverages the fact that PPG signals contain multiple physiological components related to cardiac activity, respiration, and motion. For example, Boukhechba et al. \cite{boukhechba2019actippg_27} decomposed PPG signals into cardiac components, respiration-related signals, and motion artifact components, and used these features to classify activities such as standing, walking, jogging, jumping, and sitting. Beyond single-modality approaches, several studies have explored multimodal frameworks that combine PPG with complementary sensors such as accelerometers, ECG, and electrodermal activity, which generally improve activity recognition performance \cite{mekruksavanich2022classification_55, hnoohom2023physical_154, mekruksavanich2022cnn_59, gilmore2024human}.

\subsubsection*{Others}
Beyond the primary application domains discussed above, PPG combined with deep learning has also been explored in a variety of additional healthcare scenarios. Several studies investigate systemic condition screening and severity stratification. For example, LSTM-based modeling of PPG has been used to grade autonomic nervous system dysfunction, demonstrating potential for infectious disease severity assessment in resource-limited settings \cite{tadesse2020multi_135}. Related work has extended to disease detection tasks such as sepsis detection \cite{lombardi2022classifying_57, alvarez2024lstm}, COVID-19 diagnosis \cite{lombardi2023covid_64}, and dengue severity classification \cite{karolcik2024towards}. 

PPG-driven models have also been applied in perioperative and critical care monitoring, including pain level classification \cite{lim2019deep_10}, depth of anesthesia prediction \cite{chowdhury2021deep_91}, intraoperative nociception monitoring \cite{abdel2025multimodal}, and intracranial pressure waveform reconstruction \cite{nair2024deep}. Additional applications focus on functional and physiological status assessment, such as adolescent physical fitness evaluation \cite{guo2022teenager_25}, fatigue detection \cite{liu2024medical, yu2024driver}, and neurological event identification including epileptic seizure detection \cite{yu2023artificial}. Emerging studies further explore biochemical and hematological inference tasks, including creatinine estimation \cite{sridevi2025integrated}, blood typing \cite{kavitha2025pixel}, and systemic sclerosis classification \cite{iqbal2023deep}. 

Together, these diverse yet clinically relevant applications illustrate the expanding scope of PPG-based deep learning, extending beyond traditional cardiovascular and respiratory monitoring toward broader multi-system health assessment.

\subsection*{Models}
Among the 460 papers included in this review, 283 employed CNN-based models, 89 used recurrent neural network (RNN)-based models, 96 used CRNN architectures, 38 adopted Transformer-based models, 36 used FCs, 28 used U-Net architectures, 20 used autoencoder (AE) models, 13 used GAN models, 3 used diffusion models, and 2 explored large language models (LLMs). In addition, 17 papers employed other neural network architectures, such as spiking neural networks and Mamba-based architectures. To ensure consistent statistics, the following counting rules were adopted. First, when multiple categories of deep learning models were used in the same paper, each category was counted separately. Second, for generative models, the generative framework and its backbone architecture were counted independently. For example, a U-Net-based generative model was categorized as both U-Net and CNN. Third, when a proposed model integrates multiple types of neural networks but does not clearly belong to a mainstream architecture category, each neural network component was counted individually.

As shown above, CNN-based models constitute the majority of architectures used for PPG analysis, followed by CRNN, RNN, and Transformer-based models. Several factors may explain this distribution.

\paragraph{CNN}
CNN is the most widely adopted model architecture in PPG signal analysis. Its prevalence can be attributed to several factors. First, PPG signals exhibit distinctive local morphological patterns, such as the systolic peak and the dicrotic notch. One-dimensional convolution operations are well suited for capturing such local structures, and stacking convolution and pooling layers gradually enlarges the receptive field, enabling models to capture broader temporal contexts. For many PPG-related tasks, this local pattern modeling mechanism has proven sufficient for extracting informative signal features and has demonstrated strong empirical performance \cite{moulaeifard2026machine, hong2017encase}. Second, the parameter sharing mechanism of CNNs substantially reduces the number of trainable parameters while allowing efficient parallel computation. In addition, some studies transform one-dimensional PPG signals into two-dimensional representations using signal processing techniques, which allows models to leverage mature 2D CNN architectures.

\paragraph{RNN}
RNNs and their variants are designed for sequential data and are theoretically suitable for modeling temporal dependencies in PPG signals. However, they are less frequently used as standalone models in PPG signal analysis. One reason is that the recurrent computation of RNNs requires sequential processing across time steps, which limits parallelization and typically results in higher computational costs compared with convolution-based models. In addition, PPG signals are often sampled at relatively high frequencies, producing long temporal sequences. Directly feeding long raw sequences into RNNs may lead to optimization difficulties such as gradient vanishing or exploding during backpropagation. Although gated architectures such as LSTM and GRU partially alleviate these issues, training stability and efficiency can still be challenging when modeling long PPG signal sequences.

\paragraph{CRNN}
To combine the advantages of CNNs in feature extraction and RNNs in temporal modeling, hybrid CRNN architectures have been widely adopted in PPG signal analysis. In a typical design, CNN layers first extract local morphological features from PPG signals while reducing the temporal resolution of the input sequence. The resulting feature representations are then fed into recurrent layers, such as LSTM or GRU, to capture temporal dependencies and dynamic variations over time. This design allows CNNs to efficiently capture local signal structures and compress the sequence length before applying recurrent temporal modeling, thereby improving computational efficiency while preserving temporal information.

\paragraph{Transformer}
In recent years, Transformer architectures have achieved remarkable success in fields such as natural language processing and computer vision due to their self-attention-based modeling mechanism \cite{vaswani2017attention, khan2022transformers, tucudean2024natural}. Unlike convolutional or recurrent structures, Transformers model dependencies between different positions in a sequence through attention mechanisms, enabling flexible modeling of long-range relationships. However, their application in PPG signal analysis remains relatively limited. Several factors may contribute to this phenomenon. First, Transformers typically benefit from large-scale datasets to fully exploit their representational capacity \cite{vaswani2017attention}, whereas many existing PPG datasets remain relatively small. Second, the computational complexity of the self-attention mechanism generally scales quadratically with sequence length, and PPG signals often have high sampling rates and long temporal durations, resulting in substantial computational and memory requirements. In addition, compared with vision and language domains where large pretrained models are widely available, pretrained Transformer models specifically designed for PPG signals are still limited. Nevertheless, with the increasing availability of large-scale PPG datasets and the emergence of foundation models, Transformer-based architectures may offer promising opportunities for PPG signal analysis in future research.

In addition to discriminative models, generative models have also been explored in several PPG-related tasks, including signal reconstruction, denoising, and cross-modality signal generation. Among the reviewed studies, AE-based frameworks appear relatively frequently, whereas GAN-based approaches are less commonly adopted. More recently, diffusion-based generative models have also begun to emerge in this area. In addition, the U-Net architecture is frequently used as a backbone network in generative frameworks for physiological signal modeling. Possible explanations for these observations are discussed below.

\paragraph{AE}
AE-based models are relatively commonly used in PPG-related signal reconstruction and denoising tasks. Through an encoder-decoder structure, AEs learn compact latent representations that capture important structural characteristics of PPG waveforms. This property makes them suitable for tasks such as noise suppression and artifact removal in PPG recordings. In addition, AE-based models are generally easier to train than GANs because they rely on reconstruction-based objectives rather than adversarial optimization.

\paragraph{GAN} 
GANs have also been explored for PPG-based generative tasks. However, their adoption in PPG analysis remains relatively limited. One possible reason is the inherent instability of adversarial training, which can lead to issues such as mode collapse and training divergence. In addition, PPG signals exhibit subtle waveform morphology and temporal dependencies that must be preserved during generation. Modeling these characteristics can be challenging for GAN-based frameworks, which may partly account for their comparatively limited use in PPG-related generative applications.

\paragraph{Diffusion model}
Diffusion-based generative models have recently begun to emerge in PPG-related studies. These models learn the underlying data distribution through an iterative denoising process that progressively transforms noise into structured signals. Although only a limited number of studies have applied diffusion models to PPG signals so far, preliminary results suggest that they may provide a promising framework for PPG-related generative tasks \cite{chen2025versatile, fang2025ppgflowecg}.

\paragraph{U-Net}
U-Net is a commonly used backbone architecture in PPG-related generative frameworks, particularly for signal reconstruction and denoising tasks. Originally developed for biomedical image segmentation, U-Net adopts an encoder-decoder structure with skip connections that allow low-level features from early layers to be directly combined with higher-level representations during decoding. When adapted to PPG signals using one-dimensional convolutions, this design enables models to preserve fine-grained waveform structures while learning higher-level temporal features. Consequently, U-Net has been widely adopted as the backbone architecture in several PPG generative models, including AE-based and diffusion-based frameworks.

\subsection*{Data}
Among the reviewed papers, the majority of studies relied on open-source datasets for model development. These publicly available datasets mainly originate from several representative sources, including in-hospital intensive care units (ICUs), sleep laboratories, and wearable devices such as smartwatches and wristbands. To facilitate future research, we summarize representative open-source datasets containing PPG signals in Table \ref{data}. For each dataset, key information is provided, including a brief overview, population characteristics, and the corresponding access link. In the following, we briefly introduce several representative datasets that are widely used or considered promising for future research.

\begin{itemize}
    \item The UK Biobank \cite{bycroft2018uk} is a large-scale prospective cohort study that enrolled more than 500,000 participants aged 37-73 years between 2006 and 2010 in the United Kingdom. Within this cohort, PPG signals were collected during four assessment center visits, comprising 286,819 recordings from 236,313 participants. The measurements were obtained using an infrared fingertip sensor (PulseTrace PCA2, CareFusion, USA), and the released PPG data consist of averaged beat-level pulse waveforms, each represented by 100 data points.
    
    \item The MC-MED dataset \cite{kansal2025mc} comprises 118,385 adult patient encounters recorded at the Stanford Health Care Emergency Department between September 2020 and September 2022. It includes continuous recordings of vital signs and physiological waveforms, such as PPG, ECG, and respiratory signals. In addition, the dataset provides rich clinical metadata, including patient demographics, medical histories, clinical orders, medication prescriptions, laboratory test results, imaging findings, and documented clinical outcomes.
    
    \item The Medical Information Mart for Intensive Care (MIMIC)-series Waveform Databases, including the MIMIC-II Waveform Database \cite{goldberger2000physiobank}, MIMIC-III Waveform Database \cite{johnson2016mimic}, and MIMIC-IV Waveform Database \cite{PhysioNet-mimic4wdb-0.1.0}, are commonly used ICU waveform datasets. Taking the widely used MIMIC-III Waveform Database as an example, it contains 67,830 waveform records from approximately 30,000 ICU patients. These records include digitized physiological signals such as PPG, ECG, ABP, and respiration, typically sampled at 125 Hz. In addition, when linked with the corresponding MIMIC-series Clinical Databases, the waveform data can be associated with rich clinical information for further analysis.

    \item The VitalDB dataset \cite{lee2022vitaldb} comprises 6,388 surgical cases collected at Seoul National University Hospital between 2016 and 2017. It contains high-resolution intraoperative physiological recordings from multiple anesthesia monitoring devices, including PPG, ECG, ABP, respiratory waveforms, and oxygen saturation signals. In addition, the dataset provides comprehensive perioperative clinical information extracted from the hospital electronic medical record system, such as patient demographics, surgical and anesthesia-related variables, clinical outcomes, and laboratory test results.
    
    \item The PulseDB dataset \cite{wang2023pulsedb} is a benchmark dataset designed for evaluating cuffless BP estimation methods. It consists of synchronized 10-second segments of PPG, ECG, and ABP waveforms (125 Hz) from 5,361 subjects, together with demographic information such as age. PulseDB comprises two subsets: the MIMIC-III subset (N=2,423), which contains recordings from ICU patients at Beth Israel Deaconess Medical Center between 2001 and 2012 \cite{johnson2016mimic}, and the VitalDB subset (N=2,938), which includes intraoperative recordings from surgical patients at Seoul National University Hospital in the Republic of Korea \cite{lee2022vitaldb}.

    \item The Multi-Ethnic Study of Atherosclerosis (MESA) dataset \cite{chen2015racial} is a longitudinal cohort established between 2000 and 2002 to investigate the prevalence and progression of subclinical cardiovascular disease across diverse ethnic populations. Between 2010 and 2012, 2,237 participants underwent a dedicated sleep examination that included overnight PSG recordings, in which the PPG channel was sampled at 256 Hz.

    \item The PPG-DaLiA dataset \cite{misc_ppg-dalia_495} is a multimodal dataset designed for HR estimation during daily life activities. It contains synchronized physiological and motion data recorded from 15 subjects performing eight activities, including sitting, ascending and descending stairs, table soccer, cycling, driving, lunch break, walking, and working. The dataset provides PPG signals sampled at 64 Hz from a wrist-worn device (Empatica E4), together with ECG, three-axis accelerometer, and respiration signals sampled at 700 Hz from a chest-worn device (RespiBAN Professional).
\end{itemize}

\begin{scriptsize}
\renewcommand{\arraystretch}{1.2}

\begin{longtable}{>{\raggedright\arraybackslash}m{3cm}
                  >{\raggedright\arraybackslash}m{3.2cm}
                  >{\raggedright\arraybackslash}m{6cm}
                  >{\raggedright\arraybackslash}m{4.5cm}}

\caption{Publicly available datasets for PPG research.}
\label{data} \\

\toprule
\textbf{Dataset} & \textbf{PPG Collection Device} & \textbf{Population / Overview} & \textbf{Link} \\
\midrule
\endfirsthead

\multicolumn{4}{c}{\small\itshape Table \thetable\ continued from previous page} \\
\toprule
\textbf{Dataset} & \textbf{PPG Collection Device} & \textbf{Population / Overview} & \textbf{Link} \\
\midrule
\endhead

\midrule
\multicolumn{4}{r}{\small\itshape Continued on next page} \\
\endfoot

\bottomrule
\endlastfoot

UK Biobank \cite{bycroft2018uk} &
Infrared fingertip sensor &
Large-scale prospective cohort study including more than 500,000 participants aged 37-73 years in the United Kingdom (2006-2010). &
\url{https://www.ukbiobank.ac.uk/} \\

MC-MED \cite{PhysioNet-mc-med-1.0.1} &
Bedside patient monitors &
118,385 emergency department visits from 70,545 adult patients at Stanford Health Care Emergency Department (2020-2022). &
\url{https://physionet.org/content/mc-med/1.0.1/} \\

MIMIC-III Waveform Database \cite{PhysioNet-mimic3wdb-1.0} &
Bedside patient monitors &
67,830 waveform records from approximately 30,000 ICU patients. &
\url{https://physionet.org/content/mimic3wdb/1.0/} \\

MIMIC-IV Waveform Database \cite{PhysioNet-mimic4wdb-0.1.0} &
Bedside patient monitors &
Initial release containing 200 waveform records from 198 ICU patients. &
\url{https://physionet.org/content/mimic4wdb/0.1.0/} \\

VitalDB \cite{PhysioNet-vitaldb-1.0.0} &
Operating room patient monitors &
6,388 surgical cases recorded at Seoul National University Hospital during routine or emergency operations. &
\url{https://physionet.org/content/vitaldb/1.0.0/} \\

PulseDB \cite{wang2023pulsedb} &
Patient monitors &
Aggregated dataset containing 2,423 recordings from MIMIC-III and 2,938 recordings from VitalDB for cuffless BP estimation. &
\url{https://github.com/pulselabteam/PulseDB} \\

UCI-BP \cite{kachuee2015cuff} &
Bedside patient monitors &
Subset of the MIMIC-II Waveform Database containing 12,000 simultaneous PPG and ABP records from 942 patients. &
\url{https://archive.ics.uci.edu/dataset/340/cuff+less+blood+pressure+estimation} \\

MIMIC-BP \cite{sanches2024mimic} &
Bedside patient monitors &
380 hours of synchronized ABP, PPG, and ECG recordings from 1,524 subjects extracted from MIMIC-III. &
\url{https://www.kaggle.com/datasets/tinallaksika/mimic-bp} \\

HSP \cite{sun2023human_sleep_project} &
PSG &
26,200 PSG studies from 19,492 patients collected at the Massachusetts General Hospital Sleep Laboratory. &
\url{https://bdsp.io/content/hsp/2.0/} \\

MESA \cite{chen2015racial} &
PSG &
2,237 participants underwent overnight PSG and 7-day wrist actigraphy in the MESA Sleep Exam. &
\url{https://www.sleepdata.org/datasets/mesa} \\

CFS \cite{redline1995familial} &
PSG &
Family-based sleep apnea cohort including 2,284 participants from 361 families. &
\url{https://www.sleepdata.org/datasets/cfs} \\

ABC \cite{bakker2018gastric} &
PSG &
49 adults with severe obstructive sleep apnea and class II obesity. &
\url{https://www.sleepdata.org/datasets/abc} \\

HomePAP &
PSG &
373 adult patients with moderate-to-severe obstructive sleep apnea from seven accredited sleep centers. &
\url{https://www.sleepdata.org/datasets/homepap} \\

CHAT \cite{marcus2013randomized} &
PSG &
464 children aged 5-9.9 years with mild-to-moderate obstructive sleep apnea. &
\url{https://www.sleepdata.org/datasets/chat} \\

CAP \cite{terzano2001atlas} &
PSG &
108 PSG recordings including healthy subjects and patients with multiple sleep disorders. &
\url{https://physionet.org/content/capslpdb/1.0.0/} \\

STAGES &
PSG &
Multi-center dataset including 1,500 patients evaluated for sleep disorders. &
\url{https://sleepdata.org/datasets/stages} \\

Wrist PPG During Exercise \cite{jarchi2016description} &
Shimmer 3 wrist device &
8 healthy participants recorded during walking, running, and cycling. &
\url{https://physionet.org/content/wrist/1.0.0/} \\

PPG-DaLiA \cite{misc_ppg-dalia_495} &
Empatica E4 wristband &
15 participants recorded during eight daily-life activities. &
\url{https://archive.ics.uci.edu/dataset/495/ppg+dalia} \\

BioSec.Lab PPG Dataset \cite{yadav2018evaluation} &
Fingertip pulse sensor &
University of Toronto students aged 19–35 recorded under multiple physiological conditions. &
\url{https://www.comm.utoronto.ca/~biometrics/PPG_Dataset/data_desc.html} \\

BIDMC PPG and Respiration Dataset \cite{pimentel2016toward} &
Bedside patient monitors &
53 recordings (8 minutes each) extracted from the MIMIC-II waveform database. &
\url{https://physionet.org/content/bidmc/1.0.0/} \\

CapnoBase Dataset &
Operating room patient monitors &
42 patients undergoing elective surgery under routine anesthesia. &
\url{https://peterhcharlton.github.io/RRest/capnobase_dataset.html} \\

Synthetic Dataset \cite{charlton2016assessment} &
Synthetic signal generation &
Simulated physiological signals used for respiratory rate estimation benchmarking. &
\url{https://peterhcharlton.github.io/RRest/synthetic_dataset.html} \\

Vortal Dataset \cite{charlton2016assessment} &
Pulse oximeter &
57 young and elderly participants recorded at rest and during exercise recovery. &
\url{https://peterhcharlton.github.io/RRest/vortal_dataset.html} \\

Real-World PPG Dataset &
IoT-based PPG sensor &
35 healthy participants aged 10–75; includes 1,374 training signals and 700 testing signals. &
\url{https://data.mendeley.com/datasets/yynb8t9x3d/3} \\

PPG-BP Database \cite{liang2018new} &
Pulse oximeter &
657 PPG recordings from 219 participants aged 20–89 with hypertension or diabetes. &
\url{https://figshare.com/articles/dataset/PPG-BP_Database_zip/5459299} \\

University of Queensland Vital Signs Dataset \cite{liu2012university} &
Operating room patient monitors &
PPG signals recorded during 32 surgical procedures under anesthesia. &
\url{https://outbox.eait.uq.edu.au/uqdliu3/uqvitalsignsdataset/index.html} \\

WESAD \cite{schmidt2018introducing} &
Empatica E4 wristband &
15 participants recorded during a laboratory stress and affect study. &
\url{https://archive.ics.uci.edu/dataset/465/wesad+wearable+stress+and+affect+detection} \\

UBFC-Phys Dataset \cite{sabour2021ubfc} &
Wristband device &
56 participants recorded during rest, speech, and arithmetic tasks. &
\url{https://sites.google.com/view/ybenezeth/ubfc-phys} \\

UMMC Simband Dataset \cite{bashar2019atrial} &
Simband 2 smartwatch &
35 participants with cardiac arrhythmia aged 50–91 years. &
\url{https://github.com/Cassey2016/UMass_Simband_Dataset} \\

\end{longtable}
\end{scriptsize}

\section*{Discussions}
\subsection*{PPG analysis expanded beyond traditional tasks using deep learning}
Traditionally, PPG signals have been primarily used for estimating fundamental physiological parameters such as HR and SpO$_2$ \cite{shelley2007photoplethysmography}, typically derived through signal processing algorithms based on physiological and optical principles. With the emergence of deep learning, PPG analysis has expanded substantially beyond these conventional tasks. Recent studies have demonstrated the feasibility of leveraging PPG signals for more complex applications, such as sleep staging \cite{radha2021deep_11, korkalainen2020deep_81} and ECG signal reconstruction \cite{sarkar2021cardiogan_51, chen2025versatile}. These applications rely on intricate temporal and morphological patterns that are difficult to capture using traditional feature engineering approaches \cite{islam2023revealing, han2019review}. Large-scale analyses further highlight the broader physiological relevance of PPG signals. For example, a recent phenome-wide disease detection study on the MC-MED dataset evaluated 1,468 disease phenotypes and reported meaningful discrimination (area under the receiver operating characteristic curve $\ge$ 0.70) for 307 phenotypes across 16 phecode chapters, including 230 non-circulatory conditions such as dementia, chronic kidney disease, hyperkalemia, and glaucoma \cite{nie2025anyppg}. These findings suggest that PPG signals encode rich physiological information reflecting systemic hemodynamic dynamics, while deep learning provides powerful tools for uncovering complex patterns embedded within these signals.

\subsection*{Device design and real-world validation enable wearable PPG applications}
PPG sensing has rapidly evolved from traditional clinical measurements to continuous monitoring in daily life through diverse wearable devices \cite{min2025wearable, williams2023wearable, shah2025automated, franklin2023synchronized}. This transition has substantially expanded the potential of PPG for health monitoring. However, translating wearable PPG technologies into reliable practical solutions requires careful consideration of both device design and real-world validation. Device design determines how PPG signals are acquired under different usage scenarios, whereas real-world validation ensures that these systems maintain reliable performance in clinical and everyday environments. 

\paragraph{Scenario-driven design of PPG devices.}
PPG devices are deployed across diverse application scenarios, and their designs must therefore account for scenario-specific constraints. For example, long-term monitoring emphasizes portability, comfort, and stable skin contact, motivating compact wearable forms such as rings and smartwatches \cite{elgendi2019use, sel2023continuous}. In addition to device form factors, optical sensing configurations also play an important role in determining signal quality. In particular, the optical wavelength used for PPG sensing represents a critical design consideration. The interaction between light and biological tissues is strongly wavelength dependent, as different wavelengths exhibit distinct penetration depths and absorption characteristics. For instance, shorter wavelengths such as blue light primarily interact with superficial capillary blood, whereas longer wavelengths such as red light penetrate deeper tissues and interact with arterioles and larger arteries \cite{franklin2023synchronized, liu2016multi, hossain2021quantitative}. Consequently, PPG signals acquired at different wavelengths capture hemodynamic information from different vascular layers \cite{ray2021review}. These wavelength-dependent sensing properties further influence device design; for example, green light is widely adopted in wearable PPG sensors because its shorter wavelength reduces sensitivity to deeper tissue motion while producing stronger pulsatile intensity variations associated with cardiac activity \cite{maeda2011advantages}.

\paragraph{Validation of device performance in real-world applications.}
In addition to hardware design considerations, evaluating device performance in real-world settings is essential for translating PPG-based technologies into practical health monitoring systems. Several studies have investigated the feasibility of wearable PPG devices in real-world clinical applications. For example, Chen et al.~\cite{chen2021single} demonstrated that a smartwatch-based PPG system (HUAWEI Watch GT2) can effectively screen for obstructive sleep apnea, achieving promising accuracy compared with home sleep apnea tests and PSG. Similarly, Moon et al.~\cite{moon2020validation} reported that cuffless BP estimation using the InBodyWATCH wearable device showed strong agreement with manual sphygmomanometer measurements, suggesting its potential for ambulatory BP monitoring. In addition, Selder et al.~\cite{selder2023accuracy} evaluated a standalone PPG-based AF detection algorithm integrated into consumer wrist devices and reported high sensitivity and specificity compared with reference ECG. Together, these studies highlight the growing potential of wearable PPG systems for real-world cardiovascular monitoring, while also emphasizing the need for further large-scale and longitudinal evaluations to establish their reliability and clinical utility.

\subsection*{Preprocessing strategies shape PPG model learning and generalization}
Due to the diversity of acquisition devices and real-world deployment scenarios, raw PPG signals often exhibit substantial variability, making preprocessing an essential component of deep learning-based PPG analysis \cite{pillai2025papagei, fuadah2025advances}. Common preprocessing operations, including segmentation, resampling, denoising, signal quality assessment, and data augmentation, are typically applied to organize signals into model-ready inputs, control signal quality, and improve robustness to acquisition artifacts. Importantly, these preprocessing decisions do more than standardize inputs; they fundamentally influence the information available to downstream models and therefore shape the representations learned by deep learning algorithms \cite{nolin2026foundation}. For example, segmentation strategies determine whether models capture long-range temporal dynamics or fine-grained pulse morphology, while resampling and filtering may modify the spectral characteristics of the signal. Similarly, signal quality control determines which samples are retained for training; overly strict filtering may inadvertently exclude abnormal PPG waveforms associated with disease conditions, thereby biasing the effective data distribution observed by the model. As a result, differences in preprocessing pipelines across studies may introduce dataset-specific biases and hidden distribution shifts, potentially affecting model generalization across datasets and devices. Similar concerns have been highlighted in ECG deep learning research, where inconsistent preprocessing strategies have been shown to influence model performance and hinder fair comparison between studies \cite{nolin2026foundation}. Consequently, standardized and transparently reported preprocessing protocols are essential not only for improving model robustness but also for ensuring reproducibility and enabling reliable cross-dataset evaluation in deep learning-based PPG research.

\subsection*{Model design considerations for deep learning-based PPG analysis}
When developing models for PPG analysis, several key considerations should be taken into account, including the selection of model architecture, the incorporation of physiological knowledge, the ability to address inter-individual variability through personalization, the interpretability of model predictions, and the balance between model performance and computational efficiency.

\paragraph{Model architecture}
From an architectural perspective, the model structure should align with the characteristics of PPG signals and the requirements of the target task. CNNs have been widely adopted in PPG analysis because of their ability to capture local morphological patterns in pulse waveforms, and they serve as the backbone of several existing PPG foundation models \cite{pillai2025papagei, saha2025pulse, nie2025anyppg}. In many common PPG analysis tasks, such as HR estimation, CNN-based architectures provide an effective and computationally efficient solution. In contrast, Transformer-based architectures offer stronger scalability and are particularly suitable for modeling long-range temporal dependencies or integrating multimodal inputs \cite{chen2019develop, journals/access/ShresthaM19}.

\paragraph{Physiological knowledge integration}
Beyond architectural design, incorporating domain knowledge from cardiovascular physiology can further improve model effectiveness. Prior knowledge about pulse morphology, physiological constraints, or clinically meaningful signal features can be embedded into model structures, feature representations, or training objectives, thereby improving robustness and interpretability \cite{journals/corr/abs-2005-10691, journals/bib/MiottoWWJD18}. Such knowledge-guided designs may help models focus on physiologically meaningful patterns rather than relying solely on data-driven correlations.

\paragraph{Personalization}
Personalization is another important consideration, as PPG signals often exhibit substantial inter-individual variability arising from differences in vascular properties, age, and health conditions \cite{nie2025anyppg}. Modeling strategies that incorporate subject-specific adaptation or personalized calibration may therefore improve predictive performance and clinical applicability \cite{8428413, journals/corr/abs-1909-02803, leitner2021personalized_148}. In addition, personalized modeling approaches may enhance the robustness of PPG-based systems when deployed across diverse populations and real-world environments.

\paragraph{Interpretability}
Interpretability remains a critical requirement for PPG analysis, particularly in healthcare settings where model outputs must be understandable to clinicians. Designing models that provide transparent explanations or physiologically meaningful feature attribution can facilitate clinical trust and enable deeper insights into the relationship between PPG signals and underlying physiological conditions \cite{https://doi.org/10.1002/wsbm.1548, 9852458}. Future research may further explore interpretable learning frameworks that align model predictions with clinically meaningful physiological mechanisms.

\paragraph{Balancing model performance and efficiency}
Finally, achieving an appropriate balance between model complexity and computational efficiency is essential for practical deployment \cite{hu2021model, blasch2018deep, qin2022advances}. Although highly complex models may achieve strong predictive performance, they may also introduce substantial computational overhead, which can limit their applicability in resource-constrained wearable devices or real-time monitoring scenarios. Careful consideration of such trade-offs is therefore necessary when designing models that are not only accurate but also efficient, interpretable, and suitable for real-world deployment.

\subsection*{Limited data availability, subject variability, and lack of standardization hinder PPG research progress}
Despite the growing interest in PPG-based health monitoring, several data-related challenges continue to limit research progress. First, the number of publicly available PPG datasets remains relatively small, restricting the scale and diversity of data available for model development and evaluation \cite{reiss2019deep_93, burrello2022embedding_104, loh2022application, neha2021arrhythmia}. Existing large-scale datasets, such as MIMIC and MC-MED, are primarily collected in clinical environments, and large real-world datasets acquired in daily-life settings are still scarce. Beyond data quantity, substantial inter-subject variability also presents an important challenge. Factors such as skin tone, ethnicity, lifestyle, and physiological conditions can significantly influence PPG signal characteristics \cite{Ajmal:21, 9278523, Allen_2020}. Additional variability arises from differences in sensing devices, measurement locations, and activity conditions during acquisition \cite{10.3389/fphys.2019.00198, PMID:36104356, Bent2020InvestigatingSO}. Such heterogeneity makes it difficult to develop models that generalize reliably across populations and real-world environments, highlighting the need for large-scale PPG datasets collected from diverse sources and populations. Another important issue is the lack of standardized data acquisition protocols and preprocessing procedures, which complicates the integration and comparison of datasets across studies \cite{khan2022advancement}. Furthermore, the limited availability of widely accepted benchmark datasets hinders fair and systematic evaluation of competing algorithms \cite{khan2022advancement, eerikainen2020atrial}. Together, these challenges underscore the importance of developing larger, more diverse, and better standardized PPG datasets to support robust model development and meaningful cross-study comparisons.

\subsection*{Data security and privacy considerations in PPG-based AI}
As PPG-based deep learning systems increasingly rely on large-scale physiological data, ensuring data security and privacy protection has become an essential consideration in their development and deployment \cite{miao2023wearable, sehrawat2022artificial}. PPG signals collected from wearable devices may contain sensitive health-related information, and therefore require appropriate safeguards throughout the data lifecycle, including secure storage, controlled access, anonymization, and responsible data-sharing practices. Beyond traditional health data privacy concerns, advances in machine learning introduce additional challenges. Deep learning models trained on large-scale PPG datasets may capture subtle subject-specific patterns that could potentially reveal identifiable information. In fact, studies on PPG-based biometric identification have demonstrated that PPG signals can reliably distinguish individuals \cite{biswas2019cornet_63, wang2020differences}, indicating that these signals may encode unique physiological signatures. Such findings suggest that privacy risks may extend beyond explicit personal identifiers, highlighting the need for robust data governance frameworks and ethical guidelines to ensure the responsible use of PPG data in AI-driven health monitoring systems.

\subsection*{Leveraging labeled and unlabeled data for scalable PPG modeling}
Traditional deep learning approaches for PPG analysis largely rely on supervised learning, where models are trained using labeled datasets for specific tasks. Although supervised models can achieve strong performance, their applicability is often constrained by the limited availability of high-quality labeled data. To address this limitation, recent studies have explored training paradigms that leverage both labeled and unlabeled data, including semi-supervised and self-supervised learning strategies \cite{qi2020small, van2020survey, kaul2022deep}. By exploiting intrinsic structures within PPG signals, these approaches enable models to learn transferable representations from large volumes of unlabeled data and subsequently adapt them to downstream tasks with relatively small labeled datasets \cite{radha2021deep_11, osathitporn2023rrwavenet_183, sundrani2023predicting_161, torres2020multi_136}.

Building upon these developments, foundation models have recently emerged as a promising paradigm for PPG signal analysis. Through large-scale pretraining, these models aim to learn generalizable representations that support diverse downstream tasks \cite{saha2025pulse, nie2025anyppg, pillai2025papagei, ding2025ai, chen2025gpt, abbaspourazad2023large}, often achieving stronger performance than task-specific supervised models. In practice, pretrained models can serve as powerful feature extractors or as initialization for downstream training, enabling more efficient learning when labeled data are limited. Future research may further advance this paradigm by exploring multimodal foundation models that incorporate additional physiological signals as well as textual or clinical information during pretraining \cite{pham2026pulselm}. 

\subsection*{Large language models and AI agents expand opportunities for PPG-based personalized healthcare}
Recent advances in LLMs and AI agents provide new opportunities for integrating PPG signal analysis with broader clinical and contextual information. By leveraging their ability to process unstructured medical text and perform complex reasoning, LLMs can facilitate the joint analysis of PPG signals with electronic health records, clinical notes, and other patient-related data \cite{abd2023large, zhou2023survey}. Such multimodal integration may provide richer contextual understanding of a patient's physiological state and support more informed clinical decision-making in tasks such as disease risk assessment, monitoring, and personalized treatment planning.

Beyond backend data analysis, LLM-powered agents also offer new possibilities for human-AI interaction in health monitoring systems. Integrated into wearable health platforms or mobile applications, these agents can interact with users through natural language, collect relevant health information, interpret PPG-derived indicators, and provide personalized feedback or recommendations. By bridging physiological signal analysis with conversational interfaces and clinical knowledge, LLMs and AI agents may enable more accessible, interpretable, and personalized health monitoring systems in real-world settings \cite{zhao2026ai, moritz2025coordinated}.

\section*{Conclusions}
\rev{This review surveys recent advances in deep learning-based PPG analysis across three dimensions: application tasks, model families, and data resources. Deep learning has shifted PPG analysis from handcrafted feature pipelines toward end-to-end representation learning, reducing reliance on manual feature engineering, improving task performance, and enabling modeling directly from raw signals. Despite these advances, several challenges persist, including the scarcity of large-scale, diverse, high-quality datasets, particularly in out-of-hospital settings, limited interpretability and deployment efficiency, and insufficient external and real-world validation. Future progress may benefit from balancing predictive performance with interpretability and deployment feasibility, establishing standardized and clinically representative benchmarks, and prioritizing rigorous validation under real-world conditions. In parallel, emerging paradigms such as LLMs and AI agents may further facilitate real-world adoption by enhancing the accessibility and usability of PPG-based systems. Collectively, these advances stand to accelerate the translation of PPG-based AI toward reliable tools for continuous, personalized health monitoring.}

\section*{\rev{Acknowledgments}}

\rev{\paragraph*{Ethical Approval.} As this scoping review exclusively analyzed previously published studies, it did not involve any new human participants, animal experiments, or identifiable individual-level data, and therefore did not require ethical approval.}

\rev{\paragraph*{Data Availability.} The literature summary generated for this review is available in the GitHub repository at \url{https://github.com/Ngk03/DL_PPG_Review}.}

\rev{\paragraph*{Funding.} This work was supported by the Natural Science Foundation of Beijing (QY23040), the National Natural Science Foundation of China (62102008), CCF Tencent Rhino-Bird Open Research Fund (CCF-Tencent RAGR20250108), CCF-Zhipu Large Model Innovation Fund (CCF Zhipu202414), PKU-OPPO Fund (BO202301, BO202503), Research Project of Peking University in the State Key Laboratory of Vascular Homeostasis and Remodeling (2025 SKLVHR-YCTS-02), and Beijing Municipal Science and Technology Commission (Z251100000725008).}

\rev{\paragraph*{Author Contributions.} G.N. contributed to the literature review, synthesis, and writing the original draft. J.Z. contributed to the literature review and validation. G.T., D.Z., S.G., and Q.Z. contributed to reviewing and editing the manuscript. S.H. contributed to supervision, reviewing, and editing the manuscript. All authors read and approved the final manuscript.}

\rev{\paragraph*{Competing Interests.} Shenda Hong is an Associate Editor of Health Data Science. The remaining authors declare no competing interests.}

\rev{\paragraph*{AI Tool Usage Declaration.} This manuscript was prepared with the assistance of OpenAI GPT (GPT-4 and GPT-5) and Xiaomi MiMo-V2.5 for grammar refinement and language polishing. The authors take full responsibility for the content of this publication.}

\rev{\paragraph*{Supplementary Materials.} None.}
    
\bibliographystyle{vancouver}
\bibliography{ref}

\end{document}